\documentclass[conference]{IEEEtran}
\IEEEoverridecommandlockouts
\usepackage{cite}
\usepackage{amsmath,amssymb,amsfonts}
\usepackage{algorithmic}
\usepackage{graphicx}
\usepackage{textcomp}
\usepackage{xcolor}
\def\BibTeX{{\rm B\kern-.05em{\sc i\kern-.025em b}\kern-.08em
    T\kern-.1667em\lower.7ex\hbox{E}\kern-.125emX}}

\usepackage{soul}
\usepackage{times}
\usepackage{empheq}
\usepackage{latexsym}
\usepackage{booktabs}
\usepackage{multirow}
\usepackage{subfigure}
\usepackage{tabularx}
\usepackage{color}

\begin{document}

\title{Document-aware Positional Encoding and Linguistic-guided Encoding for Abstractive Multi-document Summarization}

\author{\IEEEauthorblockN{1\textsuperscript{st} Congbo Ma}
\IEEEauthorblockA{
\textit{The University of Adelaide}\\
Adelaide, Australia \\
congbo.ma@adelaide.edu.au}
\and
\IEEEauthorblockN{2\textsuperscript{nd} Wei Emma Zhang}
\IEEEauthorblockA{
\textit{The University of Adelaide}\\
Adelaide, Australia \\
wei.e.zhang@adelaide.edu.au}
\and
\IEEEauthorblockN{3\textsuperscript{rd} Pitawelayalage Dasun Dileepa Pitawela}
\IEEEauthorblockA{
\textit{The University of Adelaide}\\
Adelaide, Australia \\
pitawelayalagedasun.pitawela@student.adelaide.edu.au}
\and
\IEEEauthorblockN{4\textsuperscript{th} Yutong Qu }
\IEEEauthorblockA{
\textit{The University of Adelaide}\\
Adelaide, Australia \\
yutong.qu@student.adelaide.edu.au}
\and
\IEEEauthorblockN{5\textsuperscript{th} Haojie Zhuang}
\IEEEauthorblockA{
\textit{The University of Adelaide}\\
Adelaide, Australia \\
haojie.zhuang@adelaide.edu.au}
\and
\IEEEauthorblockN{6\textsuperscript{th} Hu Wang}
\IEEEauthorblockA{
\textit{The University of Adelaide}\\
Adelaide, Australia \\
hu.wang@adelaide.edu.au}
}

\maketitle

\begin{abstract}

One key challenge in multi-document summarization is to capture the relations among input documents that distinguish between single document summarization (SDS) and multi-document summarization (MDS). Few existing MDS works address this issue. One effective way is to encode document positional information to assist models in capturing cross-document relations. However, existing MDS models, such as Transformer-based models, only consider token-level positional information. Moreover, these models fail to capture sentences' linguistic structure, which inevitably causes confusions in the generated summaries. Therefore, in this paper, we propose document-aware positional encoding and linguistic-guided encoding that can be fused with Transformer architecture for MDS. For document-aware positional encoding, we introduce a general protocol to guide the selection of document encoding functions. For linguistic-guided encoding, we propose to embed syntactic dependency relations into the dependency relation mask with a simple but effective non-linear encoding learner for feature learning. Extensive experiments show the proposed model can generate summaries with high quality. 
\end{abstract}


\section{Introduction}

Multi-document summarization (MDS) aims at generating fluent and informative summaries from a set of topic-related documents. Similar to single document summarization (SDS), the process of summary generation in MDS can be divided into two types: extractive summarization and abstractive summarization. Abstractive summarization requires the model to have profound natural language understanding, based on which the generated summaries could be formed by new words, phrases, or sentences that do not exist in the original documents \cite{paulus2018deep}; while extractive summarization models select existing sentences from the original documents.
Such that, in principle, summaries generated by abstractive summarization models can have higher readability, as well as conciseness, than extractive summarization models \cite{han2020abstractive}.

\begin{figure}[t]
\centering
\includegraphics[width=0.3\textwidth]{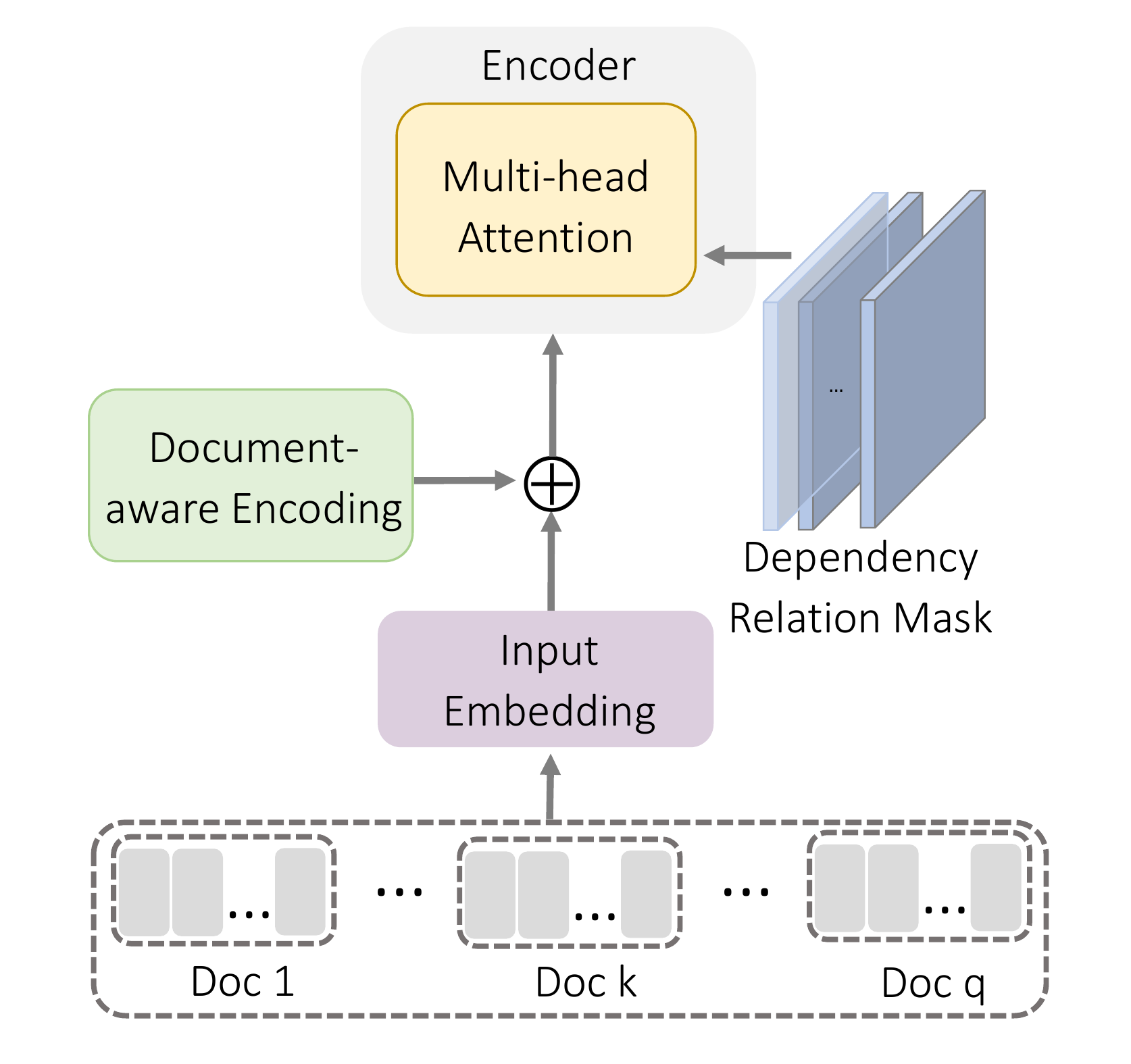}
\caption{The framework of our proposed document-aware positional encoding and linguistic-guided encoding. Document-aware positional encoding serves as part of the input of the encoder; the proposed dependency relation mask will be incorporated with multi-head attention.}
\vspace{-2mm}
\label{fig:framework}
\end{figure}

Recent years have witnessed an increasing number of neural network models applied in MDS due to the rapid improvement of computational power \cite{ma2020multi}.  Transformer is a popular one among them. It is based on a self-attention mechanism and has natural advantages for capturing cross-token relations. 
Liu et al. \cite{peter2018generwiki} proposed a Flat Transformer-based decoder-only sequence transduction to generate the Wikipedia articles. Besides Flat Transformer, Hierarchical Transformer-based models \cite{liu2019hierarchical, wei2020levegraph, pasunuru2021efficiently} utilize multiple encoders to embed the hierarchical relations 
among the source documents. To inject positional information for textual sequences, token-level positional encoding has been considered before data being pumped into encoders and decoders of these Transformer-based models. However, token-level positional encoding is not sufficient to capture document-level positional information. Missing document-level positional encoding significantly prevents models from detecting cross-document relationships. In addition to the aforementioned issue, there is another issue that obstructs the performance of Transformer-based MDS models: Transformer can explicitly compute relations between every token via a self-attention mechanism. However, this mechanism lacks explicit syntactic support that will cause the content irrelevance and deviation problem for the generated summaries \cite{hanqi2020semsum}. 
 
Dependency parsing represents the grammatical structure between each pair of words and it has been widely used in a variety of natural language processing tasks to help the model retain the syntactic structure \cite{taku2002japadepen, kai2020relational}. When it comes to document summarization, according to Hirao et al. \cite{tsutomu2004depbassen}, no matter how the word order changes from the source documents to generated summaries, the dependency structures will keep consistent in most cases. Incorporating dependency structures into summarization models is crucial to retain the correct logics from source documents.

To solve the above-mentioned two problems, in this paper, we propose an encoding mechanism combining document-aware positional encoding and linguistic-guided encoding for abstractive MDS. Figure \ref{fig:framework} illustrates a general overview of the proposed method. We construct a document-aware positional encoding protocol to guide the encoding process and the selection of document-level positional encoding functions. Like most of the Transformer-based models, we add document-aware positional encoding with the input token embedding at the bottoms of the encoder stacks. Furthermore, a novel linguistic-guided encoding is introduced to incorporate the dependency relation mask, containing pair-wise dependency relations, into the Transformer-based multi-head attention. The proposed linguistic-guided encoding method allows the model to better understand the relationship between each pair of words, and retains the correct dependency structure as well as grammatical associations when generating the summaries. We highlight our contributions as follows:

\begin{itemize}
\item We propose an effective and informative encoding mechanism to encode the multi-document positional information and the dependency structure for MDS tasks. We further propose a general protocol to guide the selection of document encoding functions.

\item We compare the proposed model with multiple competitive baselines. The results demonstrate that models equipped with the proposed encoding mechanism receive superior performances over the comparing models. We conduct an ablation study to assess the contribution of different encoding methods.

\item Extensive analysis on various settings of the document-aware positional encoding and linguistic-guided encoding are provided. These results help researchers understand the intuitiveness of the proposed model and could serve as an informative reference to the MDS research community.
\end{itemize}

\section{Related Work}
\noindent \textbf{Abstractive Multi-document Summarization.}
Abstractive MDS has been an active area of the natural language processing community in recent years. Yang et al.\cite{yang2019ht} augmented the Transformer architecture \cite{peter2018generwiki} by encoding multiple documents hierarchically. Zhang et al. \cite{jianmin2018adapneu} tried to tackle the MDS problem by utilizing a hierarchical encoder-decoder framework \cite{jiwei2015hiernu} with a \textit{PageRank} \cite{lawrence1999pagerank} based attention module. Fabbri et al. \cite{alexander2019multi} introduced an end-to-end \textit{Hierarchical maximal margin relevance-Attention Pointer-generator (Hi-MAP)} model, which expanded the existing pointer-generator network into a hierarchical fashion. It incorporates the hidden-state-based maximal margin relevance module with sentence-level representations to generate abstractive summaries. Li et al. \cite{wei2020levegraph} followed the encoder-decoder architecture with graph representations to gather rich cross-document relationships while encoding process. However, these models do not take dependency relations into account, which can assist summarization models in fetching the grammatical structure of a sentence within source documents. Song et al. \cite{kaiqiang2020jointpar} developed a shift-reduce dependency parsing system to guide summary generation by ``SHIFT" operation and pairwise dependency arc addition by ``REDUCE" operation. It transformed source sequences into summary sequences in the linearized parse tree form. Jin et al. \cite{hanqi2020semsum} constructed semantic dependency graphs by utilizing the off-the-shelf semantic dependency parser \cite{yufei2018nmsparser}. Nevertheless, little consideration paid to syntactic dependency information in the MDS area. In this paper, we propose a linguistic-guided encoding to incorporate dependency knowledge with a strong dependency learner for better attention. 

\vspace{1mm}
\noindent \textbf{Positional Encoding for Transformer.} 
Due to considering each token separately, Transformer does not keep internal sequential or order information. However, in natural language, tokens with incorrect order result in different meanings or incorrect grammar. Sequential information is crucial for language models. Vaswani et al. \cite{ashish2017attention} appended position encoding, representing the relative or absolute token position information, to the token embeddings. Sukhbaatar et al. \cite{sainbayar2019adapatten} embraced the relative position encoding assigned by a pre-defined piecewise function. Wang et al. \cite{benyou2020encoword} extended word embedding vectors to continuous word functions with absolute global positions as independent variables to model the smooth shift among sequential positions of words. 
However, these models only consider token-level positional encoding in the Transformer architecture but fail to be aware of document-level positional information, which will cause models to fail to identify different source documents. 
Therefore, in this paper, we propose to encode document-level information into positional encoding. It enables the model to perceive the token with explicit document positions for easier model optimization. Moreover, we further propose a general protocol to guide the selection of document encoding functions.

\section{Methodology}
In this work, we incorporate two types of encodings into Transformer-based abstractive MDS model: i) \textit{document-aware positional encoding} considers document positional information; ii) \textit{linguistic-guided encoding} incorporate dependency information into summarization process.  
The encodings will be introduced based on the following problem formulation and notations: given a set of $q$ documents $D=\left ( d^{1},  d^{2},..., d^{q} \right )$ on the same topic, the task of MDS is to generate a concise and informative summary $Sum$ distilling knowledge from $D$. Let $t_{i}^{k}$ denotes the $i$-$th$ token in the $k$-$th$ document $d^{k} (k = 1, 2, ... , q)$ in $D$.
$e_{i}^{k}$ represents the token embedding assigned to $t_{i}^{k}$ by the Transformer model. 

\subsection{Document-aware Positional Encoding}
\label{sec:doc_enc}
\begin{figure}[t]
\centering
\includegraphics[width=0.22\textwidth]{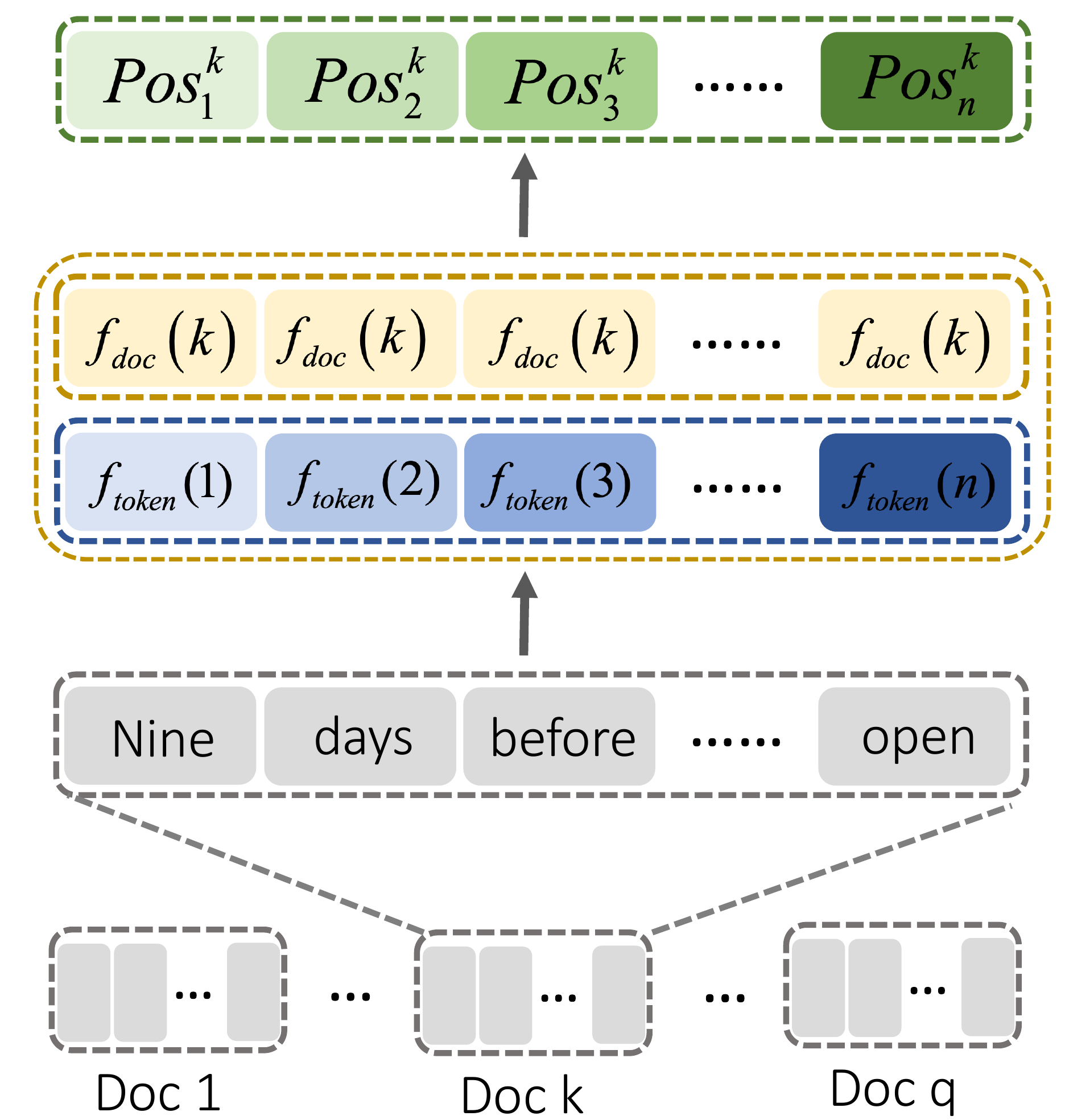}
\caption{The proposed document-aware positional encoding. It contains a document-level positional encoding and a token-level positional encoding. The selection of document positional encoding functions is according to our proposed protocol. 
}
\vspace{-2mm}
\label{fig:positional_encoding}
\end{figure}

For the token $t_{i}^{k}$ from the source documents, the token positional encoding $Pos_{token_{i}}^{k}$ and document positional encoding $Pos_{doc_{i}}^{k}$ can be represented as:

\begin{equation} \small
\begin{aligned}
& Pos_{token_{i}}^{k} = f_{token}\left ( i \right ) \\
& Pos_{doc_{i}}^{k} = f_{doc}\left ( k \right )
\end{aligned}
\end{equation}
\noindent where $f_{token}$ and $f_{doc}$ are encoding functions for token positional encoding and document positional encoding respectively. Different from the token positional embedding that considers the order of tokens, document positional embedding does not require the document order information as the order does not affect the MDS tasks.
In order to find a proper $f_{doc}$, we design a protocol with three considerations: 
(1) The encoding of each document should be unique. 
The purpose is to distinguish the documents and trace the source document for the tokens.  
(2) The values of encoding should be bonded. It will inevitably introduce large bias to certain documents if the encoding values are not bonded. (3) The values of encoding can not be remarkably larger than the value of token positional encoding. It will overwhelm the values of the token positional encoding if the document encoding values are too large, which impedes the model optimization process.
 
We adopt the $\sin$ function as document positional encoding function.
Many other functions satisfying the document positional encoding protocol.  
We discuss their performances in Section \ref{sec:sub_ablation}. 
The final positional encoding $Pos_{i}^{k}$ for token $t_{i}^{k}$ combines the token-level and document-level positional encoding by a linear combination:

\begin{equation} \small
\begin{aligned}
& Pos_{i}^{k} = \alpha Pos_{doc_{i}}^{k'} + Pos_{token_{i}}^{k}
\end{aligned}
\end{equation}
\noindent where
\begin{equation} \small
\begin{aligned}
& Pos_{doc_{i}}^{k'} = Stack\_dim_{token}(Pos_{doc_{i}}^{k})
\end{aligned}
\end{equation}

\noindent where $Stack\_dim_{token}(\cdot)$ is to repeat $Pos_{doc_{i}}^{k}$ for $dim_{token}$ times to have the same dimension with $Pos_{token_{i}}^{k}$.
Then the overall input representations to the Transformer-based model are obtained by simply adding the token embedding and its corresponding positional encoding: 
\begin{equation} \small
\begin{aligned}
& E_{i}^{k} = Pos_{i}^{k} + e_{i}^{k}
\end{aligned}
\end{equation}

Figure \ref{fig:positional_encoding} illustrates the process of proposed document-aware positional encoding. Given a set of documents (containing $q$ documents),
the document positional encoding combines with token positional encoding to form the document-aware positional encoding, which later serves as part of the input to the encoder of Transformer. 
\subsection{Linguistic-guided Encoding}
In order to retain the dependency structures and distill the generated summaries in a better manner, we propose a novel linguistic-guided encoding method to encode the informative dependency relations into the Transformer-based model. Specifically, for each sentence of source documents, an external dependency parser \cite{timothy2017deep} is adopted to extract the grammatical structure containing dependency relations between \textit{head} words and corresponding \textit{dependent} words. We construct the three-order tensor $Dep$ to place the dependency relations (the tokens discussed below are all from the same document, so the superscript $k$ is omitted). The specific dependency relations ${dep}_{ij} \in {Dep}$ can be defined as below:

\begin{equation} \small
{dep}_{ij} = 
\begin{cases}
v_{rel} & t_{i} \ominus t_{j}\\
0 & t_{i} \oslash t_{j}
\end{cases}
\end{equation}

\noindent where $v_{rel}$ $\in \mathbb{R}^{N*1}$ is the one-hot vector of dependency relations between token $t_{i}$ and $t_{j}$. There are a variety of dependency relations between paired words in dependency parsing and $N$ represents the total number of these dependencies. $t_i \ominus t_j$ indicates there is a dependency relation for $t_i$ and $t_j$, while $t_i \oslash t_j$ represents no existing dependency between the two tokens. To encode these dependency relations into the Transformer-based models, we first transfer the dependency tensor into a dependency encoding weight through a two-layers encoding function:

\begin{equation} \small
\begin{aligned}
& m_{ij} = F_{depEnc} (dep_{ij}) 
\end{aligned}
\end{equation}

\noindent where $F_{depEnc}$ contains two linear transformations and one LeakReLU non-linear mapping in between:

\begin{equation} \small
\label{eqn:depEnc-func}
   F_{depEnc}(x) = \text{Linear} \circ \text{LeakyReLU} \circ \text{Linear} (x)
\end{equation}

\noindent where $\circ$ represents the concatenation of multiple sub-functions. In general, we discover that the complexity of designing the encoding function for dependency information is crucial for model optimization.
A too-naive encoding function may lack the ability to embed the information well enough; while an encoding function with overly strong fitting abilities results in a slow training process and may cause failures in transforming the dependencies in an easy-optimizable manner.

\begin{figure}[t]
\centering
\includegraphics[width=0.45\textwidth]{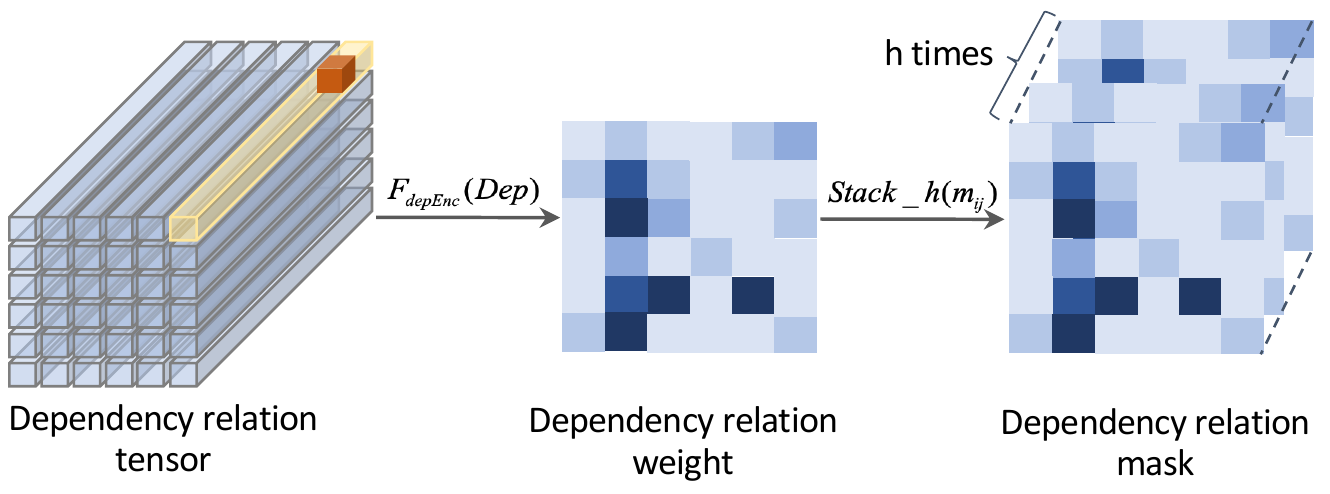}
\caption{The transformation of dependency relation mask (right) from dependency relation tensor (left).}
\vspace{-2mm}
\label{fig:dependency_encoding}
\end{figure}

Figure \ref{fig:dependency_encoding} shows the process of the transformation from dependency relation tensor ${Dep}$ to dependency relation mask $M_{ij}$. Each fiber of the dependency relation tensor represents a one-hot vector for a specific dependency relation. Only the corresponding element of the one-hot vector has the value (highlight in red). 
The dependency relation weight $m_{ij}$ is joined with the multi-head attention from source documents to generate syntactic-rich features in the following manner: 

\begin{equation} \small
MHAtt(t_{i}, t_{j}, m_{ij})=\sum_{j} \widetilde{A_{i j}} \cdot V_{j}
\end{equation}
\noindent where
\begin{equation} \small
\widetilde{A_{i j}}=M_{i j} \odot A_{i j} + A_{i j}
\end{equation}
\begin{equation} \small
A{_{ij}}=softmax\left ( \frac{{Q_i}^T K_j}{\sqrt{dim}} \right )
\end{equation}
\begin{equation} \small
M_{ij}=Stack\_h(m_{ij})
\end{equation}

\noindent where $Q_{i}, K_{j}, V_{j}\in \mathbb{R}^{h*d_k*1}$ are corresponding key, query, value for token $t_{i}$ and $t_{j}$. $dim$ is the dimension of the key, query and value. $h$ is the number of attention heads. Both $dim$ and $h$ are fixed values that we followed the original settings in Transformer. In order to fuse dependency relation weight $m_{ij}$ into dependency relation mask $M_{ij}$, function $Stack\_h(\cdot)$ is to repeat $p_{i j}$ on the dimension of head to have the same size with $Att_{i j} \in \mathbb{R}^{h*1*1}$. $\odot$ denotes the element-wise Hadamard product. Then two layer-normalization operations are applied to get the output vector of the current encoder or decoder layer for the token $t_i$.

\section{Experimental Settings}
\subsection{Datasets}

\vspace{1mm}
\noindent \textbf{Multi-News Dataset} \cite{alexander2019multi} is a large-scale English dataset containing various topics in news domain. It includes 56,216 document-summary pairs and it is further scattered with the ratio 8:1:1 for training, validation, and test respectively. Each document set contains 2 to 10 documents with a total length of 2103.49 words. The average length of the golden summaries is 263.66.

\vspace{1mm}
\noindent \textbf{Multi-XScience Dataset} \cite{yaomulti2020} is a large-scale English dataset and it contains 40,528 document-summary pairs collected from scientific articles. The task of the Multi-XScience dataset is to generate the related work section of a target scientific paper based on the abstract of the same target paper and the abstracts of the articles it refers to. The dataset contains 30,369 training, 5,066 validation and 5,093 testing data. Samples have an average input length of 778 tokens and an average length of 116 tokens on the summary.

\subsection{Implementation Details}
To have a fair comparison, we keep all the experimental settings consistent throughout all experiments. 
In our Transformer-based model, eight encoder layers and decoder layers are adopted. The Biaffine parser \cite{timothy2017deep} is used for generating dependency relations among the source documents. Our model adopts 45 dependency relations.
We use \textit{Adam} optimizer ($\beta1$=0.9 and $\beta2$=0.998) for model parameter optimization. 
The initial learning rate of the model is set to $1 \times 10^{-3}$ and 0.1 dropout rate is set for both the encoder and decoder. The trade-off hyper-parameter $\alpha$ is set to 0.1. In the training phase, the first $8 \times 10^{3}$ steps are trained for warming up and the models are trained with a multi-step learning rate reduction strategy. In the experiments, the model accumulates gradients and updates once every four iterations. The minimum and maximum lengths of the generated summaries are set to 200 and 300 words for the Multi-News dataset, while 110 and 300 words for the Multi-XScience dataset. 

\subsection{Baselines and Metrics}
\label{sec:baseline}

We compare our proposed method with the following strong baselines: \textit{LexRank} \cite{gunes2004lexrank} computes textual unit salience based on the eigenvector centrality algorithm using heuristic features in the similarity graph-based sentence representations. \textit{TextRank} \cite{rada2004textrank} leverages the graph-based ranking formula, deciding on the importance of a text unit representative within a graph built for information extraction. \textit{SummPip} \cite{zhao2020summpip} constructs sentence graphs by incorporating both linguistic knowledge and deep neural representations. \textit{Maximal Marginal Relevance (MMR)} \cite{Jaime1998mmr} combines query relevance and information novelty from source documents, benefiting summarization in reducing redundancy while remaining the most salient information. \textit{Bidirectional recurrent neural network (BRNN)} superimposes two RNNs of opposing directions on the same output according to RNN states.
\textit{Transformer} \cite{ashish2017attention} follows an encoder-decoder structure based on attention mechanism, which has been extensively utilized in a wide range of natural language processing tasks\footnote{We implement the Transformer model based on https://github.com/Alex-Fabbri/Multi-News/tree/master/code/OpenNMT-py-baselines}. \textit{CopyTransformer} restricts abstractive summarizer to copy tokens from source documents.  \textit{Pointer-Generator (PG)} \cite{seelm2017pg} equips with the coverage mechanism between the pointer network and the standard sequence-to-sequence attention model.
\textit{Hierarchical MMR-Attention Pointer-generator (Hi-MAP)} model \cite{alexander2019multi} integrates sentence representatives with hidden-state-based MMR into a standard pointer-generator network, an end-to-end model for abstract summarization. \textit{Hierarchical Transformer (HT)} \cite{liu2019hierarchical} captures relationships across multiple paragraphs via the hierarchical Transformer encoders and flat Transformer decoders\footnote{We trained the HT model on one GPU for 100,000 steps with batch-size 13,000.}.

\subsection{Automatic Evaluation Metrics}
We evaluate the models by using ROUGE scores \cite{lin2004rouge} and BERTScore \cite{tianyi2020bertscore}. Unigram and bigram overlap (ROUGE-1 and ROUGE-2 scores) are adopted to indicate the literal quality of generated summaries. ROUGE-SU score is a unigram-based co-occurrence statistic, bringing out the soft skip bigram by computing both the skip-bigram and unigram. ROUGE F1 scores are considered in our work\footnote{ The scores are computed with ROUGE-1.5.5 script with
option ``-c 95 -2 -1 -U -r 1000 -n 4 -w 1.2 -a -m"}. 
BERTScore is an automatic language evaluation metric for text generation based on contextual token embeddings of the pre-trained BERT \cite{jacob2019bert}. We mark ROUGE-1, ROUGE-2, ROUGE-SU and BERTScore as ``R-1, ``R-2", ``R-SU" and ``BS" in this paper.

\section{Experimental Results}

\begin{table} [t] \small
\centering
\caption{Performance Comparison on the Multi-News dataset. We rerun all the baseline models under the same settings. ``CopyTrans" represents CopyTransformer. The best results for each column are in bold.
} \label{tab:Multinews}
\setlength{\tabcolsep}{3mm}{
\begin{tabular}{l|cccc}
\hline
Models          & R-1   & R-2   & R-SU  & BS \\ \hline
LexRank         & 37.92 & 13.10  & 12.51 & 0.83 \\  
TextRank       & 39.02 & 14.54 & 13.08 & 0.83 \\
SummPip         & 42.29  & 13.29   & 16.16 &0.84  \\
MMR             & 42.12 & 13.19  & 15.63  & 0.84  \\
\hline
BRNN            & 38.36 & 13.55 & 14.65 & 0.83 \\ 
Transformer     & 25.82 & 5.84  & 6.91  & 0.80 \\ 
CopyTrans & 42.98 & 14.48 & 16.91 & 0.84 \\  
PG              & 34.13 & 11.01 & 11.58 &  0.83  \\ 
Hi-MAP          & 42.98 & 14.85 & 16.93 &0.83 \\ 
HT              & 36.09 & 12.64 & 12.55 & 0.84 \\
\hline
Ours           & \textbf{44.35} & \textbf{15.04} & \textbf{17.97} & \textbf{0.85}     \\ \hline
\end{tabular}}
\vspace{-2mm}
\end{table}

\begin{table} [t]\small
\label{XScience}
\centering
\caption{Performance Comparison on the Multi-XScience dataset. We rerun all the baseline models under the same settings. ``CopyTrans" represents CopyTransformer. The best results for each column are in bold.}
\setlength{\tabcolsep}{3mm}{
\begin{tabular}{l|cccc}
\hline
Models          & R-1                  & R-2                  & R-SU                 & BS            \\
\hline
LexRank         & \textbf{31.31}                & 5.85                 & 9.13      & 0.83                         \\  
TextRank        & 31.15                & 5.71                 & 9.07      & \textbf{0.84}                           \\  
SummPip         &29.66  &5.54 &8.11 & 0.82 \\
MMR             & 30.04                & 4.46                 & 8.15                 & 0.83                \\ 
\hline
BRNN            & 27.95                & 5.78                 & 8.43                 & 0.83               \\ 
Transformer     & 28.34  &4.99 &8.21        &0.82                \\ 
CopyTrans  & 26.92  &4.92   &7.50  &0.83               \\
PG              &30.30  &5.02  &9.04  &\textbf{0.84}  \\
Hi-MAP          & 30.41                & 5.85                 & 9.13                 & 0.81                \\ 
HT               &25.31 &4.23& 6.64 & 0.83 \\
\hline
Ours            & 30.93 & \textbf{6.06} & \textbf{9.57} & \textbf{0.84}  \\
\hline
\end{tabular}}
\vspace{-2mm}
\end{table}

\subsection{Overall Performance}
In this section, we compare our proposed model with several strong baselines and list the comparison results in Table \ref{tab:Multinews} (Multi-News) and Table  \uppercase\expandafter{\romannumeral2} (Multi-XScience).
The results of our proposed model on the Multi-News dataset show the best overall results on both ROUGE scores and BERTScore. To give a fair comparison, we rerun all the baseline models. It is observed that our model performs particularly well on R-SU than other models. It gains 1.06 improvement to the second best, \textit{Hi-MAP}. Given that R-SU takes more skip-bigram plus unigram-based co-occurrence statistics into account, it contains additional comprehensive information to evaluate the models. The BERTScore on different models shows relative marginal differences. However, our proposed model still achieves the best among all the evaluate models, which indicates our proposed model can generate high-quality summaries in a semantic level. 
We also evaluate our proposed models based on the Multi-XScience datasets. Comparing the \textit{Transformer} baseline models and our model with document-aware positional encoding and linguistic guided encoding, we observe that these two encodings help to improve the performance by 2.59 on R-1, 1.07 on R-2 and 1.36 on R-SU. The results on the Multi-XScience dataset show that our model performs better than most of the models. Our proposed model does not achieve the best results on all evaluation metrics because the proposed model is based on the Transformer models which are dataset sensitive. This means the Transformer-basd models do not always work well on all the MDS datasets. This phenomenon can also be found in the paper \cite{zhao2020summpip, pasunuru2021efficiently} and \cite{han2020abstractive}. In these paper, the Transformer-based model (\textit{CopyTransformer}) shows poor results on DUC-2004 dataset\footnote{http://duc.nist.gov} although it works well on the Multi-News dataset. A potential reason is Multi-XScience and DUC-2004 datasets have higher novel n-grams score than Multi-News dataset \cite{alexander2019multi, yaomulti2020}. For example, paper \cite{yaomulti2020} reported that the proportion novel of unigrams/bigrams/trigrams/4-grams in the golden summaries of the Multi-News dataset is 17.76/57.10/75.71/82.30, which are much lower than that of Multi-XScience dataset (42.33/81.75/94.57/97.62). The Transformer models may not work very well on datasets with higher novel n-gram scores.

\begin{table}[t] \small
\centering
\caption{Ablation Study of Our Model on Multi-News and Multi-XScience Dataset.  ``doc-pos en" and ``depen en" stand for document-aware positional encoding and linguistic-guided encoding.}
\label{tab:ablation_study}
\setlength{\tabcolsep}{3mm}{
\begin{tabular}{l|l|ccc}
\hline
Dataset & Model Variants & R-1            & R-2            & R-SU           \\ \hline
Multi-  & w/o doc-pos en & 44.16          & 15.06 & 17.74          \\
News    & w/o depen en   & 43.73          & 14.86          & 17.37          \\
        & Full Model     & 44.35 & 15.04          & 17.97 \\ \hline
Multi- & w/o doc-pos en & 28.81          & 5.53           & 8.56           \\
XScience & w/o depen en   & 29.69          & 5.62           & 8.86           \\
        & Full Models    & 30.93          & 6.06           & 9.57           \\ \hline
\end{tabular}
}
\vspace{-2mm}
\end{table}

\subsection{Ablation Study}
\label{sec:sub_ablation}

To better understand the contribution of document-aware positional encoding and linguistic-guided encoding techniques to overall model performance individually, we conduct an ablation study on the proposed model on both Multi-News and Multi-XScience datasets. Table \ref{tab:ablation_study} presents the results. The experiments confirm that the proposed two encoding methods perform considerably better than the model without them.  This is due to (1) document-aware positional encoding has the capability of capturing cross-document information in MDS; (2) with linguistic-guided encoding, dependency relations within the source documents are well preserved, enabling the summarization model to effectively learn a much more faithful syntactic structure than that working on the model without it.

\subsection{Encoding Strategies}

In addition to the model performance evaluation, we report our findings on different encoding functions and the ways to incorporate the encoding. 

\vspace{1mm}
\noindent \textbf{(1) Document-aware Positional Encoding Strategies.} We evaluate the contribution of different document positional encoding functions. All these functions satisfy the proposed protocol described in Section \ref{sec:doc_enc}. The experience results are shown in the upper part of Table \ref{tab:analysis}.  $\sin$ function helps the MDS model achieve the best ROUGE score and the combination of $\sin$ and $\cos$ produce similar results. However,  $\cos$ function greatly reduces the model performance.
The reason could be related to the document number in a document set of Multi-News dataset. Most of the document sets contain two documents in the Multi-News dataset. When applying $\cos$ on two documents, the value differences for the two encodings is smaller than what the $\sin$ function provides, which means $\cos$ has less distinguishing ability than $\sin$. This may result in lower model performance for MDS tasks. Additionally, we also try to adjust $\alpha$ in Equation (2). Results are shown in the lower part of Table \ref{tab:analysis}. We test the model performance on validation set when $\alpha =$ 0.1, 0.5, 1 and observe model perform best when $\alpha = 0.1$. Therefore, we fix this hyper-parameter to 0.1 and report the final results on the test set.

\begin{table} [t] \small
\centering
\caption{Performance of our model using different document positional encoding strategies. The strategies include different encoding functions (upper) and different document positional encoding weights(lower). iter(A, B) means to use functions A and B alternately. Values obtained from the validation set based on the Multi-News dataset.}
\setlength{\tabcolsep}{3mm}{
\begin{tabular}{l|ccc}
\hline
Models                     & R-1   & R-2   & R-SU  \\
\hline
$\sin(x)$                     &43.80	&14.74	&17.59 \\
$\cos(x)$                     & 42.82 & 14.49 & 16.71 \\
$iter(\sin(x), \cos(x))$               & 43.56 & 14.43 & 17.38 \\
$iter(\sin(0.1x), \cos(0.1x))$         & 43.66 & 14.52 & 17. 47   \\
\hline
$\alpha$= 0.1      &44.11	&14.81	&17.74 \\
$\alpha$= 0.5      &43.68	&14.54	&17.45 \\
$\alpha$= 1      &43.80	&14.74	&17.59 \\ 
\hline
\end{tabular}}
\vspace{-2mm}
\label{tab:analysis}
\end{table}

\vspace{1mm}
\noindent \textbf{(2) Document-aware Positional Encoding Protocol.} To verify the proposed three considerations of document encoding functions, we select some other functions except $sin$ and $cos$, and the functions are not satisfy the conditions proposed in \ref{sec:doc_enc} for experiments. We also randomly assign values to document positional encoding to verify the effectiveness of our chosen function. The results are shown in Table \ref{tab:encoding_protocol}. We observe that the performance are not well when (1) the document positional encoding of each document is the same (SameEncoding); (2) the values of document positional encoding are not bonded ($y=x$, $y=2x$, $y=5x$, $y=10x$); and (3) the values of document positional encoding are remarkable larger than the values of token positional encoding ($y=10x$); (4) randomly assign values to document positional encoding (Random).

\begin{table} [t] \small
\centering
\caption{Performance of models with functions that do not meet the document positional encoding protocol. Values obtained from the validation set based on the Multi-News dataset.}
\setlength{\tabcolsep}{3mm}{
\begin{tabular}{l|ccc}
\hline
Models                     & R-1   & R-2   & R-SU  \\
\hline
SameEncoding     & 42.82	& 14.28	& 16.63 \\
$y=x$      &42.25	&14.08	&16.25		\\
$y=2x$      &42.57	&14.06	&16.60		\\ 
$y=5x$      &40.56	&12.06	&15.13		\\ 
$y=10x$     &38.94	&11.50	&14.24		\\ 
Random    &43.19	&14.67	&16.87		 \\ 
\hline
\end{tabular}
}
\vspace{-2mm}
\label{tab:encoding_protocol}
\end{table}

\begin{table} \small
\centering
\caption{Performance of our model based on different linguistic-guided encoding methods. Values obtained from the validation set based on the Multi-News dataset.}
\label{tab:analysis_dep}
\setlength{\tabcolsep}{3mm}{
\begin{tabular}{l|ccc}
\hline
Models                 & R-1   & R-2   & R-SU  \\
\hline
Arithmetic sequence        & 43.71 &14.54 &17.43 \\
Arithmetic sequence (core) & 43.79 &14.57 &17.47 \\
Arithmetic sequence (root) & 43.89 &14.64 &17.55 \\
One-hot (one layer)    & 43.15 &14.40 &17.03 \\ 
One-hot ($F_{depEnc}$) &44.11	&14.81	&17.74 \\
Added on values         & 42.41 &13.89 &16.47 \\
\hline
\end{tabular}}
\vspace{-2mm}
\end{table}

\vspace{1mm}
\noindent \textbf{(3) Linguistic-guided Encoding Strategies.} There are 45 dependency relations existing in the Biaffine parser. Some dependency relations have a great influence on the generated summaries; and vice versa. This section discusses how to encode these various relations into multi-head attention mechanism by considering their importance. The performance of different linguistic-guided encoding methods is shown in Table \ref{tab:analysis_dep}. The importance of dependency relations in the first three methods are manually set and the following two are automatically learned. ``Arithmetic sequence" represents a sequence with the values of $1, N-1/N, N-2/N, ..., 1/N$, which means the dependency relations at the top of the list have a larger weight. $N$ denotes the number of dependency relations in total. The sequence of dependency relation list in the first methods is constructed according to the sequence of the occurrence of dependency relations in the source documents. We select the top-8 dependents from the official core dependents of clausal predicates\footnote{https://universaldependencies.org/docs/en/dep/} to build the relation lists for ``Arithmetic sequence (core)". ``Arithmetic sequence (root)" is to assign the largest weight to the root word since the dependency relation ``root" is proven to be the most important token in the syntax dependency tree \cite{kai2020relational}. ``One-hot (one layer)" means the one-hot representation of dependencies with only one linear transformation between the dependency relation tensor and the dependency relation mask. The ``One-hot (one layer)" model performs substantially poorer than the one-hot encoding model with non-linear function $F_{depEnc}$. It is because non-linearity enlarges the learning capability of encoding functions significantly. The``One-hot ($F_{depEnc}$)" represents our final model. The $F_{depEnc}$ function can outperform all arithmetic sequence models since it delegates the construction of dependency relations to a non-linear learner. It enables the model to learn the importance of gradient descent directly. From another point of view, besides the addition of the linguistic-guided encoding on keys and queries within self-attention, we also tried to add the encoding on values. However, model performance dropped greatly. We hypothesize the reason is that keys and queries are adopted to calculate attention, but values are the final receptors of attention. Small changes in values will have a large influence on the model optimization process.

\subsection{Human Evaluation}

\begin{table} [t]\small
\centering
\caption{Human evaluation results on the Multi-News dataset. The best results for each column are in bold. ``CopyTrans" represents CopyTransformer.} 
\label{tab: human-evaluation}
\setlength{\tabcolsep}{1mm}{
\begin{tabular}{l|ccc}
\hline
Models          & Fluency & Informativeness & Conciseness  \\ \hline
Transformer     & 2.50 & 1.97 & 2.50    \\
CopyTrans & 2.60 & 2.60 & 2.83     \\
Hi-MAP          & 3.07 & 2.87 & 2.97    \\
\hline
Ours            & \textbf{3.13} & \textbf{3.10} & \textbf{3.20}     \\
\hline
\end{tabular}}
\vspace{-2mm}
\end{table}

\begin{table*}[htp]
\centering
\caption{Generated summaries of different models given the same source documents. ``CopyTrans" represents CopyTransformer. Different colors represent different thought groups.} 
\label{tab: summary-examples}
\begin{tabularx}{\textwidth}{|l|X|}
\hline 
Source Docs &  \textcolor{red} {a nine-year-old boy from los banos has completed quite the journey on tuesday , as he swam through the san francisco bay , all the way to alcatraz island and back .} \textcolor{green} {james savage is hoping to be the youngest swimmer on record to make the swim . the title is currently held by a 10-year-old .} it ' s been quite an emotion day in san francisco , but it ended with smiles and cheers as james walked on shore , after hours in the bay . james began the swim at 7 : 00 a.m. , and it took him a little over two hours to complete the feat . he struggled for a big out in the open water , about 30 minutes into the swim . james said the waves were hitting him , and the current was too strong , making it difficult for him to fight them . in the end , however , with help from his coach and a promise from his father , james found his second wind , and was off . \textcolor{blue} {" i wanted to turn back and they kept yelling , ' i believe i you you can do it ' , " said james . " we kind of pushed each other . finally got him going . smooth sailing all the way }......\textcolor{red} {james savage , a 9-year-old boy from los banos , calif}......\\ 
\hline
Transformer & \textcolor{red} {james savage has been swimming off the san francisco bay area since he was 14 years old ,} \textcolor{orange} {but he ' s not the only one who ' s been in the water for more than a century . " i ' ve never seen anything like this before , " he tells nbc bay area . " it ' s a great thing to do . i ' m not going to be able to do it . "} he says he ' ll be swimming in the bay of san francisco ' s bay in san francisco when he ' d swim to the beach , and he ' s \textlangle blank\textrangle \textlangle blank\textrangle \textlangle blank\textrangle ......\\
\hline
CopyTrans & \textcolor{green} {james savage , a 9-year-old boy who completed the tiring trek last year , was able to set a record as the youngest swimmer to make the journey to the island and back on june 14 , reports nbc bay area} ...... \textcolor{orange} {james tells ksee . " he ' s not going to do anything about it , " he says , adding that his second wind was off the rails . " it ' s really hard to get to the bottom of the road , " savage says of his father . " it was just a slap on the wrist . it was a slap in the water .} it ' s \textlangle blank\textrangle \textlangle blank\textrangle \textlangle blank\textrangle ......\\
\hline
Hi-Map & \textcolor{green} {james savage is hoping to be the youngest swimmer on record to make the swim to the former prison , reports the san francisco chronicle .} \textcolor{blue} {" i wanted to turn back and they kept yelling , ' i believe i you you can do it ' , " james told ksee . " we kind of pushed each other . finally got him going . smooth sailing all the way , " he says .} " i \textlangle blank\textrangle \textlangle blank\textrangle \textlangle blank\textrangle ...... \\
\hline
Ours  & \textcolor{red} {james savage , a 9-year-old boy from los banos , calif. , has completed the swim at 7 : \textlangle unk\textrangle and 57 pounds ,} \textcolor{green} {was able to set a record as the youngest swimmer to make the journey to the alcatraz island and back on june 14 , nbc news reports.} \textcolor{blue} {" i wanted to turn back and they kept yelling , ' i believe i you can do it , ' " james tells ksee . " we kind of pushed each other . finally got him going . smooth sailing all the way , "} ...... \\
\hline
\end{tabularx}
\vspace{-5mm}
\end{table*}

Apart from automatic evaluation, we conduct a human evaluation to assess the quality of the generated summaries on three aspects: \textbf{text fluency} checks whether the summary is natural, well-formed, and both syntactically and semantically correct; \textbf{conciseness} assesses whether the summary is concise and without repeated or useless information; \textbf{informativeness} examines whether the summary keeps the salient information from the source documents. We randomly sample 10 examples from the Multi-News dataset \cite{alexander2019multi}. Three experienced researcher are invited to score summaries (from 4 models) on the above aspects. The score range is 1-5 (1 means very bad; 5 means very good). The final scores for each model are averaged across different examples and raters. The results are listed in Table \ref{tab: human-evaluation}. The text fluency score of our model is 3.13, which is higher than 2.50 of \textit{Transformer}, 2.60 of \textit{CopyTransformer}, and 3.07 of \textit{Hi-Map}, which means the summaries generated by our model are more natural and well-formed. In terms of the score of informativeness, our model achieves 3.10 and is higher than the second-best model (\textit{Hi-Map}) by 0.23, indicating our model is better at capturing the most important information from different sources. Moreover, the generated summaries by our model are more concise and better at reducing redundant information, which could be concluded by the conciseness score.

\subsection{Case Study}

Table \ref{tab: summary-examples} presents the generated summaries from four models: \textit{Transformer}, \textit{CopyTransformer}, \textit{Hi-Map}, and our models. In this example, the \textit{Transformer} model only captures ``james savage has been swimming off the san francisco bay area" (in red) but takes the age wrong. It should be 9 in fact. Besides, \textit{Transformer} model also generates something that are not supported in the source document (in orange). For the \textit{CopyTransformer}, the salient information (in green) is in the generated summary. However, this model also outputs unsupported text (in orange). The \textit{Hi-Map} model misses some key information (e.g. the red highlight in the source document).
In contrast, the summary generated by our proposed model keeps the significant information and shows content consistent with the source documents. It could demonstrate that our model equipped with the proposed informative encoding mechanism could generate summaries more accurately than the other comparing models.

\section{Conclusion}

In this paper, we propose to incorporate document-aware positional encoding and linguistic-guided encoding for abstractive multi-document summarization. We conduct extensive experiments on two benchmark datasets and the results demonstrate the superior performance of the proposed two encoding methods. The analysis of various settings of the document-aware positional encoding and linguistic-guided encoding can help researchers understand the intuitiveness of the proposed model and could serve as an informative reference to the MDS research community. In the future, we would like to explore different ways to capture cross-document relations to further improve the quality of generated summaries.

\bibliography{IEEEexample}
\bibliographystyle{IEEEtran}

\end{document}